\documentclass[conference]{IEEEtran}
\IEEEoverridecommandlockouts
\usepackage{cite}
\usepackage{amsmath,amssymb,amsfonts}
\usepackage{algorithmic}
\usepackage{graphicx}
\usepackage{textcomp}
\usepackage{xcolor}
\usepackage{float}
\usepackage{booktabs}
\usepackage{amssymb}
\usepackage{color}
\usepackage{subfigure}
\usepackage{multirow}
\usepackage{array}
\usepackage{hyperref}
\usepackage{pdfpages}
\usepackage{tabularx}

\def\BibTeX{{\rm B\kern-.05em{\sc i\kern-.025em b}\kern-.08em
    T\kern-.1667em\lower.7ex\hbox{E}\kern-.125emX}}
\begin{document}

\title{ARIC: An Activity Recognition Dataset in Classroom Surveillance Images
\thanks{ ${}^{\dag}$ Corresponding author (lfxu@uestc.edu.cn)} 
}

\author{
Linfeng Xu$^{\dag}$, Fanman Meng, Qingbo Wu, Lili Pan, Heqian Qiu, Lanxiao Wang, \\
Kailong Chen, Kanglei Geng, Yilei Qian, Haojie Wang, Shuchang Zhou, Shimou Ling,\\
Zejia Liu, Nanlin Chen, Yingjie Xu, Shaoxu Cheng, Bowen Tan, Ziyong Xu, Hongliang Li\\
    \textit{School of Information and Communication Engineering} \\
    \textit{University of Electronic Science and Technology of China, Chengdu, China}
}
% \author{
%     \IEEEauthorblockN{Yilei Qian$^{a*}$, Kanglei Geng$^{a*}$, Kailong Chen$^{a}$, Shaoxu Cheng$^{a}$, Linfeng Xu$^{a\dag}$, Hongliang Li$^{a}$, Fanman Meng$^{a}$, Qingbo Wu$^{a}$}
%     \IEEEauthorblockA{$^a$ School of Information and Communication Engineering,University of Electronic Science and Technology of China, Chengdu, China}
%     \IEEEauthorblockA{\{zhangsan\}@XXX.com, \{lisi, wangwu\}@XXX.edu.cn}
% }

\maketitle

\begin{abstract}
The application of activity recognition in the ``AI + Education" field is gaining increasing attention. However, current work mainly focuses on the recognition of activities in manually captured videos and a limited number of activity types, with little attention given to recognizing activities in surveillance images from real classrooms. Activity recognition in classroom surveillance images faces multiple challenges, such as class imbalance and high activity similarity. To address this gap, we constructed a novel multimodal dataset focused on classroom surveillance image activity recognition called ARIC (Activity Recognition In Classroom). The ARIC dataset has advantages of multiple perspectives, 32 activity categories, three modalities, and real-world classroom scenarios. In addition to the general activity recognition tasks, we also provide settings for continual learning and few-shot continual learning. We hope that the ARIC dataset can act as a facilitator for future analysis and research for open teaching scenarios. You can download preliminary data from \href{https://ivipclab.github.io/publication_ARIC/ARIC}{https://ivipclab.github.io/publication\_ARIC/ARIC.}
\end{abstract}


\section{Introduction}
\label{Introduction}
In recent years, activity recognition has gained increasing attention as a significant application of AI in classroom settings. However, existing studies\cite{jisi2021newclass,sharma2024starclass} have primarily focused on the recognition of a limited number of activities, and the data collected are often manually captured videos rather than classroom surveillance images. Activity recognition in classroom surveillance images faces multiple challenges, including class imbalance, high activity similarity, and privacy protection. To fill this gap, we constructed the ARIC (Activity Recognition In Classroom) dataset, specifically designed for activity recognition in classroom surveillance images. This dataset offers a rich variety of activity types, provides multi-perspective surveillance images, and is sourced from real classroom surveillance videos. However, the ARIC dataset also presents several challenges: 1) an imbalanced distribution of activity categories with significant differences in sample sizes; 2) high similarity between samples of different categories, which can lead to confusion; 3) features extracted by a shallow network to protect privacy, increasing recognition difficulty; and 4) settings for continual learning aiming at the continuous analysis of activities in open teaching scenarios.

\section{ARIC-Dataset}
\label{ARIC-Dataset}

\begin{table*}[ht]
    \centering
    \caption{List of Activities Categories in ARIC with Their Descriptions}
    \begin{tabular}{|c|l|p{11cm}|}
        \hline
        \textbf{Index} & \textbf{Label name} & \textbf{Description} \\
        \hline
        0  & Listening             & Student listening \\
        1  & Reading               & Reading book \\
        2  & Using\_phone          & Using phone \\
        3  & Using\_pad            & Using pad \\
        4  & Using\_computers      & Using computer \\
        5  & Scratching\_head      & Scratching head \\
        6  & Writing               & Using a pen to write, including writing on a pad using a capacitive pen. \\
        7  & Talking               & Students talking each other specially \\
        8  & Standing              & Standing \\
        9  & Sleeping              & Sleeping \\
        10 & Teaching              & The teacher is standing and lecturing to the students. \\
        11 & Yawning               & Yawning \\
        12 & Walking               & Walking \\
        13 & Relaxing              & Relaxing movements, such as stretching, neck rotation, and arm stretching. \\
        14 & Analyzing             & Analysis, specifically referring to the teacher pointing at the blackboard or PowerPoint to analyze content. \\
        15 & Taking\_bottle        & Holding a water cup/water bottle. \\
        16 & Gathering\_up\_bag    & Packing up the backpack. \\
        17 & Drinking              & Drinking water/beverage, with obvious drinking actions. \\
        18 & Taking\_photos        & Taking photos with electronic devices. \\
        19 & Listening\_to\_music  & Wearing headphones is considered as listening to music. \\
        20 & Discussing            & Discussing, specifically referring to students and teachers discussing. \\
        21 & Setting\_equipment    & Adjusting equipment, such as staff adjusting the cameras and other devices for recording the dataset. \\
        22 & Taking\_bag           & Holding a bag. \\
        23 & Blackboard\_writing   & Writing on the blackboard. \\
        24 & Blackboard\_wiping    & Erasing the blackboard. \\
        25 & Taking\_off           & Taking off clothes. \\
        26 & Student\_demonstrating & Students presenting/reports in front of the podium or classroom. \\
        27 & Eating                & Eating. \\
        28 & Reviewing             & The teacher sitting below, evaluating students' presentations/reports. \\
        29 & Hands\_up             & Raising hand. \\
        30 & Speaking              & Students standing up to speak. \\
        31 & Picking\_up\_computers & Packing up the computer. \\
        \hline
    \end{tabular}
    \label{tab:classroom_activities}
\end{table*}

\subsection{Data Collection}
\label{ssec:Data Collection}
\begin{figure}[htbp]
    \centering
    \subfigure[front]{\includegraphics[width=0.4\hsize, height=0.25\hsize]{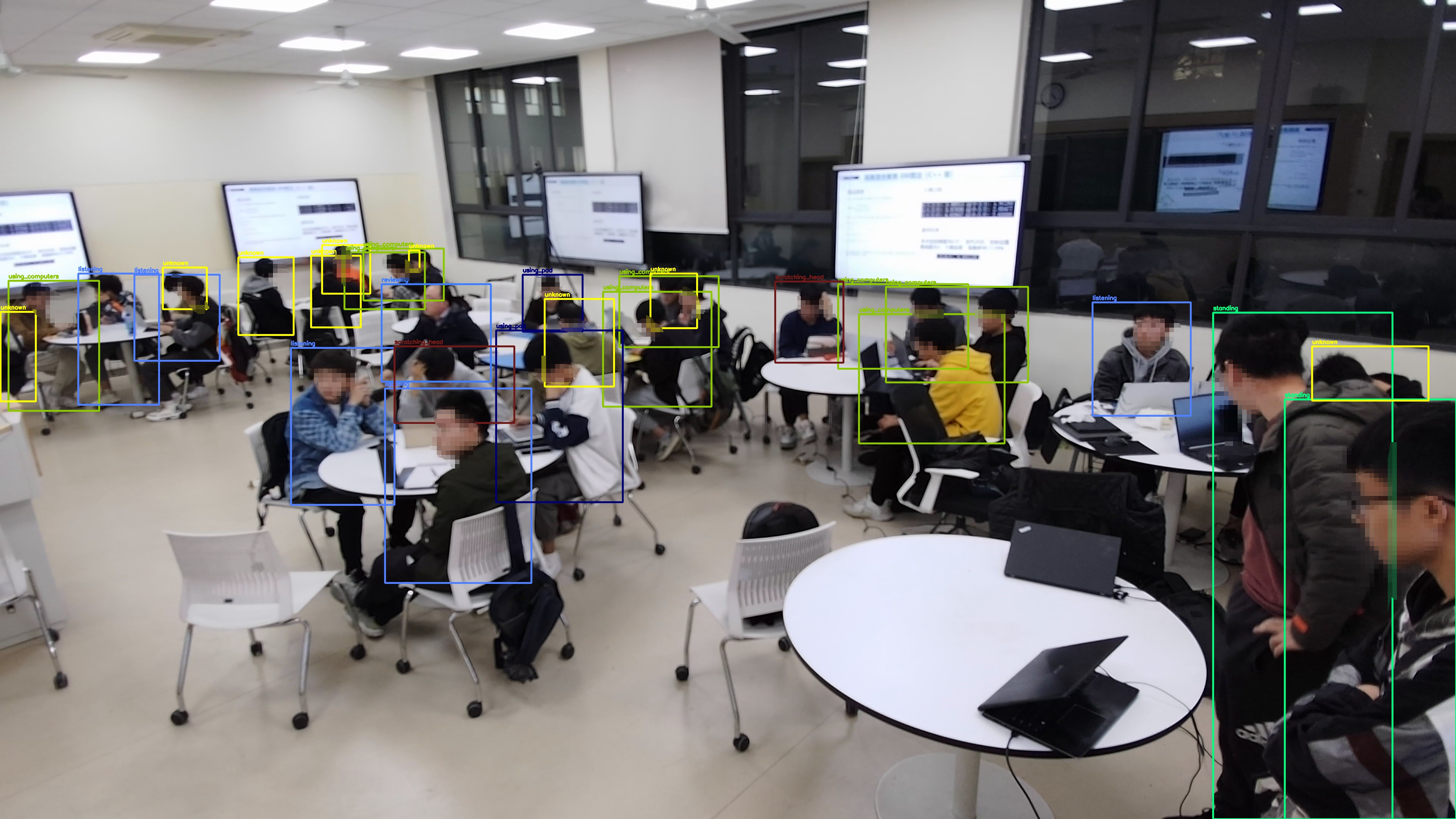}\label{data_1}}
    \subfigure[middle]{\includegraphics[width=0.4\hsize, height=0.25\hsize]{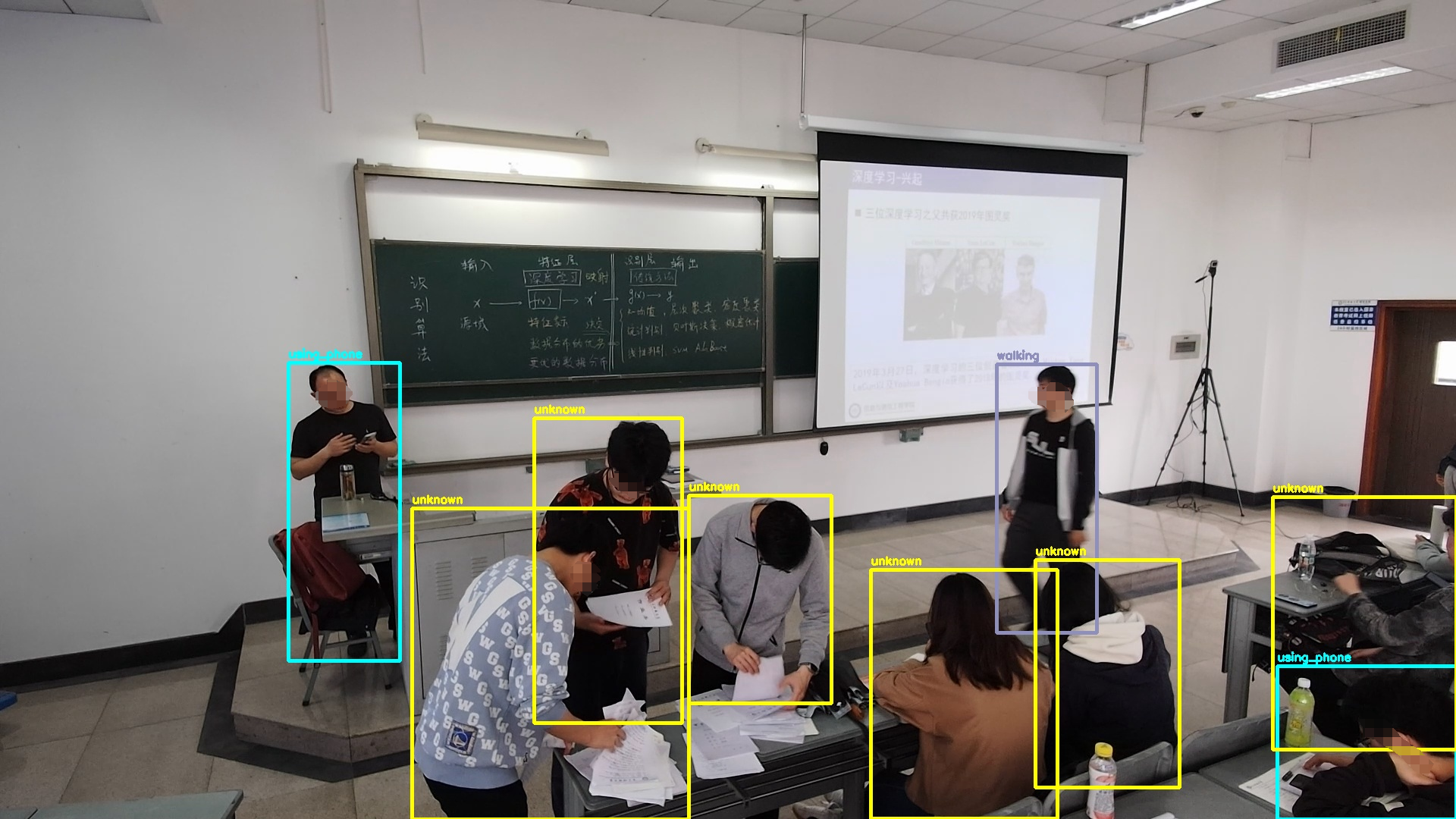}\label{data_2}}
    \subfigure[rear]{\includegraphics[width=0.4\hsize, height=0.25\hsize]{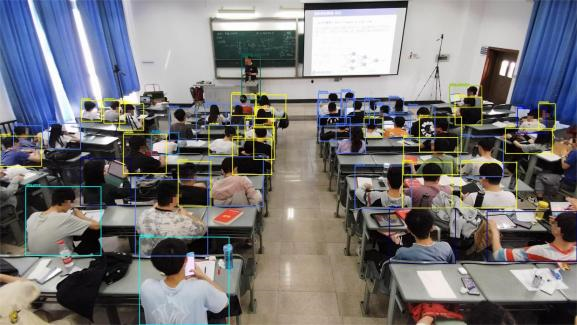}\label{data_3}}
    \caption{\small{Monitoring Images from Different Perspectives}
    \label{dataset}}
\end{figure}

To collect data for activity recognition in classroom scenarios, we captured raw surveillance video from smart classrooms during regular class sessions. This setup recorded a range of activities performed by students and teachers in both large and small classroom environments from multiple angles. High-quality samples were obtained using 4K HD cameras mounted on tripods and extension rods, providing clear, detailed recordings.For more robust feature extraction in subsequent recognition tasks, we positioned cameras strategically at the front, middle, and rear of the classroom, as illustrated in Fig.\ref{dataset}. This multi-angle setup captured varied perspectives, minimizing recognition errors caused by movement or changes in position. Such an approach ensures stable and accurate data collection, contributing to a more reliable classroom quality assessment system.

\subsection{Dataset Overview}
\label{ssec:Dataset Overview}
The ARIC is a brand-new and challenging dataset based on real classroom surveillance scenarios. The dataset includes three modalities: image, audio, and text. First, we extracted image frames from the collected HD videos and annotated the behaviors of individual instances of students and teachers to form the image modality. Next, we extracted 10-second audio clips (5 seconds before and after each image) to create the audio modality, ensuring that a complete sentence from the teacher appears in the audio. Lastly, for the text modality, we used the open-source large model InternVL\cite{chen2024internvl}, setting the prompt as “Please describe the image in detail” to generate descriptive text for each image. This text provides an overall description of the scene environment as well as detailed depictions of the behaviors of students and teachers, serving as the text modality for each image.

\begin{figure}[htbp]
    \centering
    \includegraphics[width=\linewidth]{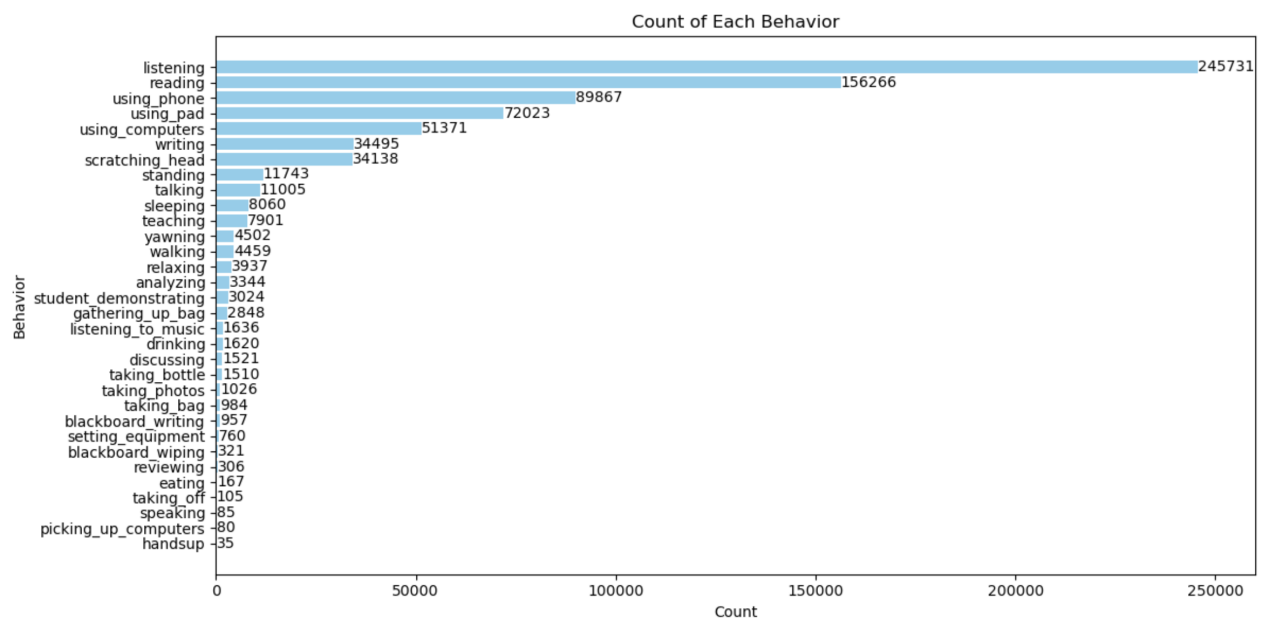}
    \vspace{-1em}
    \caption{\small{Sample Distribution of the 32 Activity Categories}}
    \label{dis}
\end{figure}

The ARIC dataset consists of 36,453 surveillance images, covering 32 types of classroom activities, including listening to a lecture, reading, using a phone, using a pad, using a computer, scratching head, writing, talking, standing, sleeping, teaching, yawning, walking, relaxing, analyzing, holding a bottle, packing a backpack, drinking water, taking photos, listening to music, discussing, debugging equipment, holding a backpack, writing on the blackboard, erasing the blackboard, taking off a coat, student presentation, eating, reviewing, raising a hand, speaking, and picking up a computer, with descriptions of each behavior provided in Table 1. Compared to the classroom activity recognition datasets in \cite{choi2015conceptual,nguyen2022new}, our dataset offers an extremely rich variety of behaviors, covering almost all possible activities that may occur in a classroom. However, the distribution of samples across these 32 activities is highly imbalanced. For example, common activities such as listening to a lecture, reading, and using a phone have tens of thousands of samples, while less common activities like speaking, eating, and raising a hand may have only a few dozen samples, as shown in Fig.\ref{dis}. This significant long-tail effect presents a major challenge for our dataset.

\subsection{Privacy Protection}
\label{ssec:privacy protection}
To protect the privacy of individuals appearing in the images and to avoid releasing the original images, we used shallow layers of pre-trained models to convert the original images into feature data. Considering the need for backbone models in the field of continual learning, we selected three commonly used pre-trained models: ResNet50\cite{he2016deep(resnet)}, ViT\cite{dosovitskiy2020image(vit)}, and CLIP-ViT\cite{radford2021learning(clipvit)}. For ResNet50, the instance images pass through the 7x7 conv1 layer, and then through the 3x3 max pooling layer, yielding output feature dimensions of [1, 64, 56, 56]. This portion constitutes the released features. For the ViT model, the released features are the patch embeddings and position embeddings obtained after inputting the instance images. The dimensions of these features are [1, 196, 768]. For the ClipViT model, two types of features are released. One is the patch embeddings and position embeddings obtained from the instance images, with output dimensions of [1, 50, 768]. The other is the embeddings obtained from the text input, with dimensions of [1, 5, 512]. The text input content is "Someone is [DOING]".

\begin{figure}[htp]
    \centering
    \includegraphics[width=\linewidth]{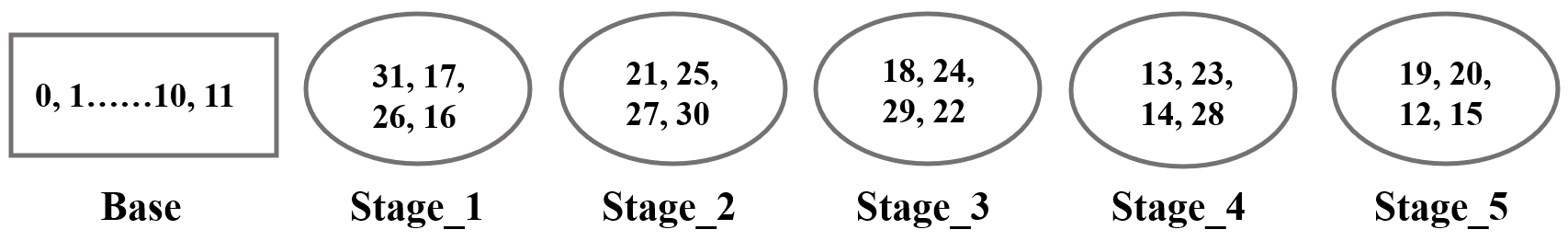}
    \vspace{-1em}
    \caption{\small{A Setting for Continual Learning Setting}}
    \label{cil}
\end{figure}

\begin{figure}[htp]
    \centering
    \includegraphics[width=\linewidth]{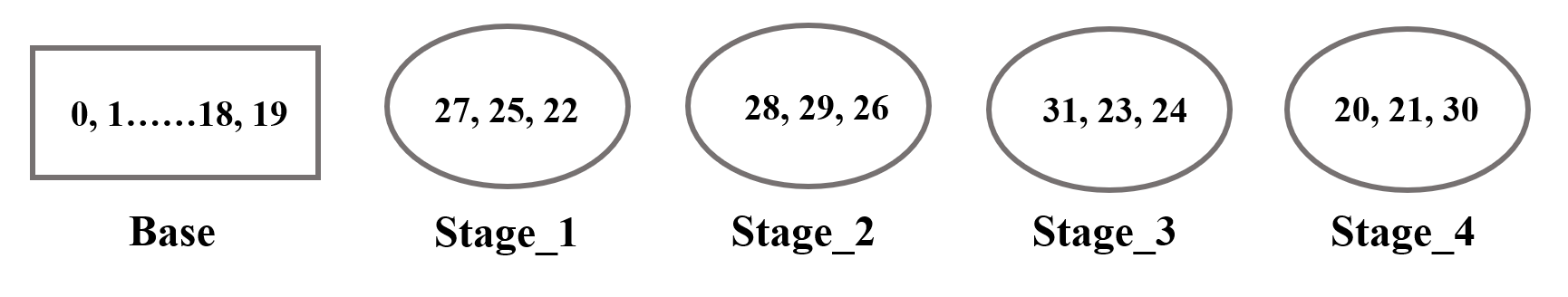}
    \vspace{-1em}
    \caption{\small{A Setting for Few-Shot Continual Learning}}
    \label{fscil}
\end{figure}

\subsection{Settings for Continual Learning}
\label{ssec:Settings for CIL}

We also pre-defined reasonable incremental learning settings within the dataset to standardize experiments across the dataset. This was done for two key reasons: (1) In the base phase, a few categories with a large number of samples are provided to ensure strong initial training. The remaining categories are then randomly and as evenly as possible distributed across different incremental steps to mimic real-world scenarios where new categories emerge over time. (2) Alternatively, the categories are arranged in descending order by the number of samples and then allocated to different incremental phases based on this order, ensuring a gradual introduction of less-represented classes. This division helps maintain balance in the learning process and simulates more realistic incremental learning settings. The specific partitioning schemes will be represented using the formula: $\boldsymbol{B}+\boldsymbol{S}\times \boldsymbol{N}$. Here, $\boldsymbol{B}$ represents the number of base classes, $\boldsymbol{S}$ represents the number of incremental steps, and $\boldsymbol{N}$ represents the number of categories in each incremental step. For example, $8+6\times4$ means there are $8$ base categories, $6$ incremental steps, and $4$ categories in each incremental step.

For continual learning, we defined two incremental modes: $8+6\times4$ and $12+5\times4$, each with two settings of different category loading sequences. For example, Fig. \ref{cil} shows one of these incremental settings. Categories with more samples were selected for the base phase to ensure strong performance during the initial training. The remaining categories are randomly and evenly distributed across the incremental phases to simulate real-world scenarios where new categories emerge over time. The number of base classes was determined by referencing other incremental learning datasets to maintain consistency with standard continual learning research practices. More flexible settings can be found in the Readme.md.

\subsection{Settings for Few-Shot Continual Learning}
\label{ssec:Settings for FSCIL}

For few-shot continual learning, we also defined two incremental modes: $16+4\times4$ and $20+4\times3$, similar to continual learning, with two settings of different category loading sequences for each mode to accommodate different experimental needs. A setting of $20+4\times3$ incremental mode is shown in Fig. \ref{fscil}, where the base phase consists of categories $0-19$, and the loading sequences for the incremental steps are as follows: Step 1: $[27, 25, 22]$, Step 2: $[28, 29, 26]$, Step 3: $[31, 23, 24]$, and Step 4: $[20, 21, 30]$. The base phase includes categories with more samples, allowing the model to learn foundational knowledge from a larger data pool. Incremental categories are introduced from less-represented classes, simulating the few-shot continual learning scenario. The number of base classes was selected based on commonly used datasets in few-shot continual learning research to ensure standardization across similar tasks and to enable fair experimental comparisons. More flexible settings can be found in the Readme.md.

\subsection{The Supplement of ARIC}
Since the ARIC dataset only provides features for individual instances and lacks global field-of-view information from the original images, we have compiled and released ARIC\_supplement to bridge this gap.

In ARIC\_supplement, we provide the shallow features of the original images after resizing them to (640, 640) and passing them through three types of pre-trained networks as we mentioned before. Additionally, we include the initial annotation JSON file, which contains the xyxy coordinates of different instances within each image.

The ARIC dataset and ARIC\_supplement can be downloaded, and more detailed information can be obtained by the link: \href{https://ivipclab.github.io/publication_ARIC/ARIC}{https://ivipclab.github.io/publication\_ARIC/ARIC.}

\newpage
\bibliographystyle{IEEEtran}
\bibliography{ARIC}

\end{document}